\documentclass[12pt]{article}
\usepackage{fontspec}

\usepackage{geometry}
\usepackage{setspace}
\usepackage{caption}
\usepackage{booktabs}
\usepackage{threeparttable}

\usepackage{amsmath}
\usepackage{mathtools}
\usepackage{amssymb} 
\usepackage{amsfonts}
\usepackage{graphicx}

\usepackage{xcolor}

\definecolor{R1}{RGB}{35,63,113}

\usepackage[doi=false,isbn=false,url=false,eprint=false,date=year,style=nature]{biblatex}
\usepackage{url}
\usepackage{hyperref}
\hypersetup{colorlinks=true,
            citecolor=blue,
            linkcolor=blue,
            urlcolor=blue,
            breaklinks=true
}

\usepackage{authblk}

\usepackage{lineno}

\title{AI has its own style in mathematical proving}

\author
{Xiaokai Li$^{1}$, Ziyu Zhou$^{2}$, Qibin Cheng$^{3}$, Yangcheng Wu$^{1}$, An Zeng$^{1\ast}$, and Ying Fan$^{1\ast}$
\\

$^{1}$School of Systems Science, Beijing Normal University, Beijing 100875, China

$^{2}$Microsoft Research Asia, Beijing 100080, China

$^{3}$School of Mathematics, Renmin University of China, Beijing 100872, China

$^{\ast}$Corresponding authors; Emails: anzeng@bnu.edu.cn; yfan@bnu.edu.cn
}

\date{}

\let\oldequation\equation
\let\oldendequation\endequation

\renewenvironment{equation}
  {\linenomathNonumbers\oldequation}
  {\oldendequation\endlinenomath}

\begin{document}
\captionsetup[figure]{font={stretch=1.2},labelfont={bf},labelformat={default},labelsep=space,name={Fig.}}
\captionsetup[table]{font={stretch=1.2},labelfont={bf},labelsep=space,name={Tab.}}

\maketitle

\section*{\centering Abstract}

As large language models increasingly generate mathematical proofs, whether they develop a proving style distinct from that of human mathematicians remains an open question.
This question is particularly pressing as mathematicians increasingly voice concerns that proofs generated by large language models are difficult to interpret.
Here we ask three large language models to independently reprove all 465 propositions of Euclid's \textit{Elements} and compare their citation choices against the human-proof baseline through Jaccard similarity and complex network analysis.
We find that AI-generated proofs differ from human proofs, and this difference is stable and reproducible: mean pairwise Jaccard similarity among AI runs (${\sim}$0.68) is nearly twice that between AI and human proofs (${\sim}$0.37), and cross-run classification consistency is also high.
By examining citation structure in detail, we identify the concrete signature of this difference: approximately 40\% of AI citations point to foundational elements (definitions, common notions, and postulates), compared with only 13\% in human proofs; even among theorem-to-theorem citations, AI tends to cite more distant results.
Roughly half of AI's extra citations trace to the upstream dependencies of the human proof.
Together, these findings suggest that the AI proving style is a tendency to construct arguments on its own from foundational elements rather than directly citing later, more established results, which we term ``building from the ground up''.
Our findings offer a new perspective on how AI systems organise and invoke mathematical knowledge, with implications for understanding the interpretability of AI-generated proofs.

\clearpage


\newrefsection

\section*{Introduction}

In 2026, large language models achieved a series of breakthroughs in mathematical proof that went well beyond earlier results.
Beyond solving competition problems\supercite{deepseek2025r1,luong-etal-2025-towards,balunovicMathArenaEvaluatingLLMs2025,zhangProposingSolvingOlympiad2026}, AI began to engage with open research questions: in May 2026, an OpenAI model disproved the Erdős unit distance conjecture\supercite{alon2026unitdistance}; in September 2026, OpenAI announced a 166-page proof addressing a formulation of the Navier--Stokes existence and smoothness problem\supercite{openai2026navierstokes}.
Yet as AI-generated proofs have moved from competition exercises to frontier mathematics, the initial enthusiasm among mathematicians has given way to caution.
Some mathematicians argue that AI-generated proofs are difficult to interpret\supercite{brumfiel2026nstokes,friedman2026anxiety}, dwell on trivial details\supercite{taoMathematicsAgeAI2026}, and are ``not written for humans''\supercite{science2026nstokes,moradifamWhatRemainsHuman2026}.
In September 2026, twenty-five Fields Medalists signed an open letter cautioning against treating problem-solving performance as an end in itself, arguing that ``solving problems is only a tool and proxy for achieving the primary goal of conceptual understanding and insight''\supercite{tao2026misalignment}.

These shifts suggest that the focus has moved, in part, from ``how hard a problem can AI prove'' to ``how does AI's proof differ from a human's''.
Given that large language models' ability to produce mathematical proofs is now well established, what interests us more is whether AI develops its own proving style.
The difficulty of interpreting AI's proofs may itself be a manifestation of this distinct style.
Some observational studies have explored differences between the characteristics of AI and human proofs\supercite{10.1145/3750038,aspertiThinkingMachinesMathematical2026,trinhSolvingOlympiadGeometry2024,hubert2025alphaproof,chenComprehensiveAnatomyHuman2026,liangStrategyExecutabilityMathematical2026a,dingAIHumanApproaches2026}.
AI-generated proofs have been found to be verbose\supercite{trinhSolvingOlympiadGeometry2024,hubert2025alphaproof}, to dwell repeatedly on local details\supercite{chenComprehensiveAnatomyHuman2026}, to favor procedural and algebraic strategies over geometric and structural insight\supercite{liangStrategyExecutabilityMathematical2026a}, and to emphasize problem resolution and cross-disciplinary integration over methodological exposition\supercite{dingAIHumanApproaches2026}.

Yet these observational studies capture only a range of manifestations of differences, leaving open whether these manifestations reflect deeper, systematic regularities in how AI constructs mathematical proofs.
Addressing this question requires systematically examining how AI draws on mathematical knowledge in its proofs and whether these choices are reproducible across independent attempts and models.
Moreover, proving style may be reflected not only in individual proofs but also in the structure of the deductive system that emerges from them.
In an axiomatic system, mathematical results are established sequentially, with each new proposition building on foundational axioms and previously established results.
These dependencies collectively form a structure that captures how a body of mathematical knowledge is constructed.
Whether AI and human mathematicians produce systematically different structures of mathematical dependence remains insufficiently understood.

To investigate these questions, we use Euclid's \textit{Elements} as a testbed for comparing AI-generated and human proofs.
The \textit{Elements} provides a self-contained axiomatic system: 465 propositions spanning plane geometry, solid geometry, and number theory, progressing from the construction of an equilateral triangle to the Pythagorean theorem and the classification of the five regular solids.
Heath's classical English translation supplies a complete proof text and a structured citation record for every proposition, providing an authoritative human baseline\supercite{wolframStructureEuclidsElements2020}.
These properties make the \textit{Elements} particularly well suited for studying proving style---both at the level of individual proofs and at the level of the deductive system as a whole.

Here, we ask large language models to reprove all 465 propositions of Euclid's \textit{Elements} independently, citing only elements available within the deductive system.
Five runs across three models---three with Claude Opus 4.6, one with Claude Fable 5.1, and one with GPT-6 Astra---test the consistency of the results.
We compare the resulting proofs against Heath's at the level of individual citation sets and, through complex network methods, at the level of global network structure.
Our results show that AI proving differs systematically from human proving: citation similarity among AI runs is nearly twice that between AI and human proofs, and this pattern is stable across independent runs and models.
AI-human divergence becomes more likely for propositions associated with greater proof complexity.
The defining characteristic of the AI proving style is a tendency to build logical chains from foundational tools rather than directly citing later, more established results---what we term ``building from the ground up''.
This style remains robust to changes in prompting.
These results provide a systematic and concrete characterization of AI's distinctive proving style, helping mathematicians understand AI-generated proofs and coexist more effectively with AI in the age of AI-driven mathematics.

\section*{Experimental design overview}

We asked three large language models to independently prove all 465 propositions (i.e., theorems) of Euclid's \textit{Elements}: Claude Opus 4.6 (Anthropic), Claude Fable 5.1 (Anthropic), and GPT-6 Astra (OpenAI).
For Claude Opus 4.6 (with reasoning effort set to ``high''), we conducted three independent runs (A, B, and C).
The 465 propositions were proved sequentially in their original order, so that each proof could cite only postulates, common notions, definitions, and propositions already established.
For each proposition, the model received only the proposition statement itself and these citable resources; it had no access to human proof text, citation set, or any output from other proving attempts.

This was achieved through a workflow-and-skill architecture within the Claude Code platform.
An orchestrating workflow spawned a dedicated sub-agent for each proposition in parallel; each sub-agent operated in its own isolated context, receiving only the inputs described above.
The proving protocol was encapsulated in a reusable skill (``euclid-proof'') that instructed the model to produce a step-by-step proof together with a USED list recording every citation actually invoked.
In practice, these operations were executed book by book (the \textit{Elements} comprises thirteen books in total).
After each book's proofs were completed, a separate post-processing step compared the AI's citation set for each proposition against Heath's using the Jaccard similarity coefficient, producing the full AI proof text and a CSV table of citation sets for that book.

To characterise how each AI proof relates to the human proof, we built a separate automated pipeline---again driven by an AI judge---that classified every AI proof into one of four categories:
P (perfect match), where the citation set is identical to Heath's; G (granularity difference), where the proof follows the same logical path but cites at a finer or coarser level of detail---for example, explicitly invoking postulates or common notions that Heath leaves implicit; D (path difference), where the proof constitutes a valid argument via a genuinely different route; and E (error), where the proof contains an identifiable logical defect.
Classification was likewise performed through a parallel workflow, with a dedicated sub-agent classifying each proposition independently according to a structured decision tree; the detailed classification protocol is described in Methods.
In addition, all geometric propositions underwent a separate geometric consistency audit that checked for spatial configuration contradictions—such as inconsistent point positions, angle decomposition errors, or circle containment conflicts—reclassifying as E where contradictions were found.

\begin{figure}[htbp]
  \centering
  \includegraphics[width=0.87\textwidth]{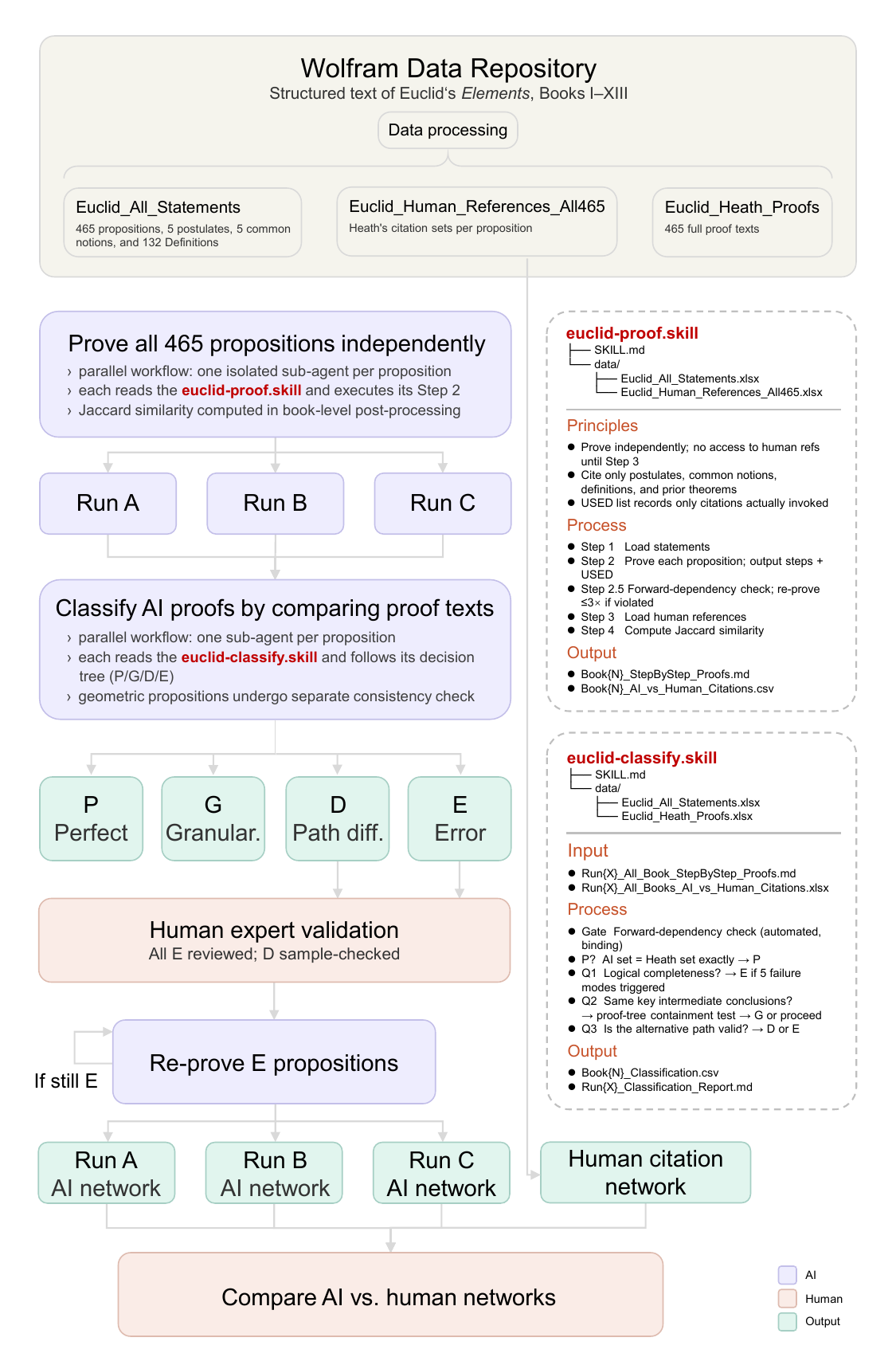}
\caption{\textbf{Experimental pipeline.}
{The left panel shows the main pipeline for the three Claude Opus 4.6 runs, from data extraction through independent proving, automated classification (P/G/D/E), expert validation, error re-proving, and citation network construction.
The right panel details the proving and classification protocols.}
}
\label{figure1}
\end{figure}

We verified the accuracy of the AI check through expert validation: two human experts reviewed all E-class proofs and a sample of D-class proofs, correcting misclassifications where necessary.
The expert review confirmed that the AI judge's classifications were largely accurate, with only a small number of corrections (see Methods).
Every confirmed error was then re-proved by the AI under the same constraints until all errors were resolved.
A flowchart of the entire experimental process is provided in Fig. \ref{figure1}.
This process yielded classification results at three stages---after AI check, after expert validation, and after re-proving---summarised in Tab. \ref{tab1}.
The final result contains zero E-class proofs in all runs;
the dominant category across all three runs is G, accounting for roughly 68.2\% of propositions, while D accounts for about 27.2\% and P about 4.6\%.
In the Results sections that follow, unless otherwise noted, all analyses involving classification results are based on the final classifications after re-proving.

\begin{table}[ht]
\centering
\caption{\textbf{Classification results across three validation stages.}}
\label{tab1}
\begin{threeparttable}
\begin{tabular}{lccccc}
\toprule
 & P & G & D & E & Err\% \\
\midrule
\addlinespace[6pt]
\multicolumn{6}{l}{\textbf{After AI check}} \\[4pt]
Run A & 22 & 317 & 115 & 11 (5) & 2.37 \\
Run B & 18 & 320 & 119 &  8 (3) & 1.72 \\
Run C & 24 & 295 & 136 & 10 (3) & 2.15 \\[4pt]
\midrule
\addlinespace[6pt]
\multicolumn{6}{l}{\textbf{After expert validation}} \\[4pt]
Run A & 22 & 319 & 115 & 9 & 1.94 \\
Run B & 18 & 320 & 120 & 7 & 1.51 \\
Run C & 24 & 295 & 136 & 10 & 2.15 \\[4pt]
\midrule
\addlinespace[6pt]
\multicolumn{6}{l}{\textbf{After re-prove}} \\[4pt]
Run A & 22 & 324 & 119 & 0 & 0 \\
Run B & 18 & 325 & 122 & 0 & 0 \\
Run C & 24 & 303 & 138 & 0 & 0 \\[4pt]
\midrule
\addlinespace[6pt]
\multicolumn{6}{l}{\textbf{Cross-model replication (after AI check)}} \\[4pt]
Fable 5.1 & 28 & 403 & 34 & 0 & 0 \\
GPT-6 Astra & 9 & 295 & 161 & 0 & 0 \\[4pt]
\bottomrule
\end{tabular}
\begin{tablenotes}[flushleft]\small
\item\textbf{Note:} E parenthetical indicates geo-FAIL errors identified from D-class proofs. P remains constant across all stages. $N = 465$ for each run. Runs A--C use Claude Opus 4.6; Fable 5.1 and GPT-6 Astra each performed one run under the same constraints.
\end{tablenotes}
\end{threeparttable}
\end{table}

To assess whether our results are specific to a single model or reflect a more general tendency of large language models, we repeated the entire experiment with anothor two models: Claude Fable 5.1 and GPT-6 Astra, both with reasoning effort set to ``xhigh''.
Each model proved all 465 propositions under the same constraints (only one run) and underwent the same classification pipeline.
The results are also summarised in Tab. \ref{tab1}.
They show that both newer, more capable models avoided errors entirely, yet their proofs exhibited different category distributions: Fable 5.1 produced fewer D-class proofs (aligning more closely with the human proof), while GPT-6 Astra produced even more D-class proofs than Opus (diverging further from it).

Despite this, we found robust commonalities across AI proofs.
We measured citation similarity both among AI runs and between AI and human proofs; the comparison reveals a proving style distinct from the human one, and cross-run consistency confirms its stability (Results, section 2).
We found divergence correlates with proposition complexity (Results, section 3).
To characterise this style structurally, we constructed citation networks for both AI and human proofs, in which nodes represent propositions, definitions, common notions, or postulates and edges represent citations invoked during proving (see Methods)\supercite{viteriEpistemicPhaseTransitions2022a,liSocietalCitationsUndermine2025,liNetworkStructureMathlib2026,kurganTheoremGraphBridgingFormal2026a}.
Comparison of network topology identifies citation patterns that distinguish the AI style from the human one (Results, section 4).
We further confirmed that these patterns remain robust across different prompts through additional experiments (see Supplementary Text 1).

\section*{Results}

\subsection*{Basic Statistics}

\begin{figure}[htbp]
  \centering
  \includegraphics[width=\textwidth]{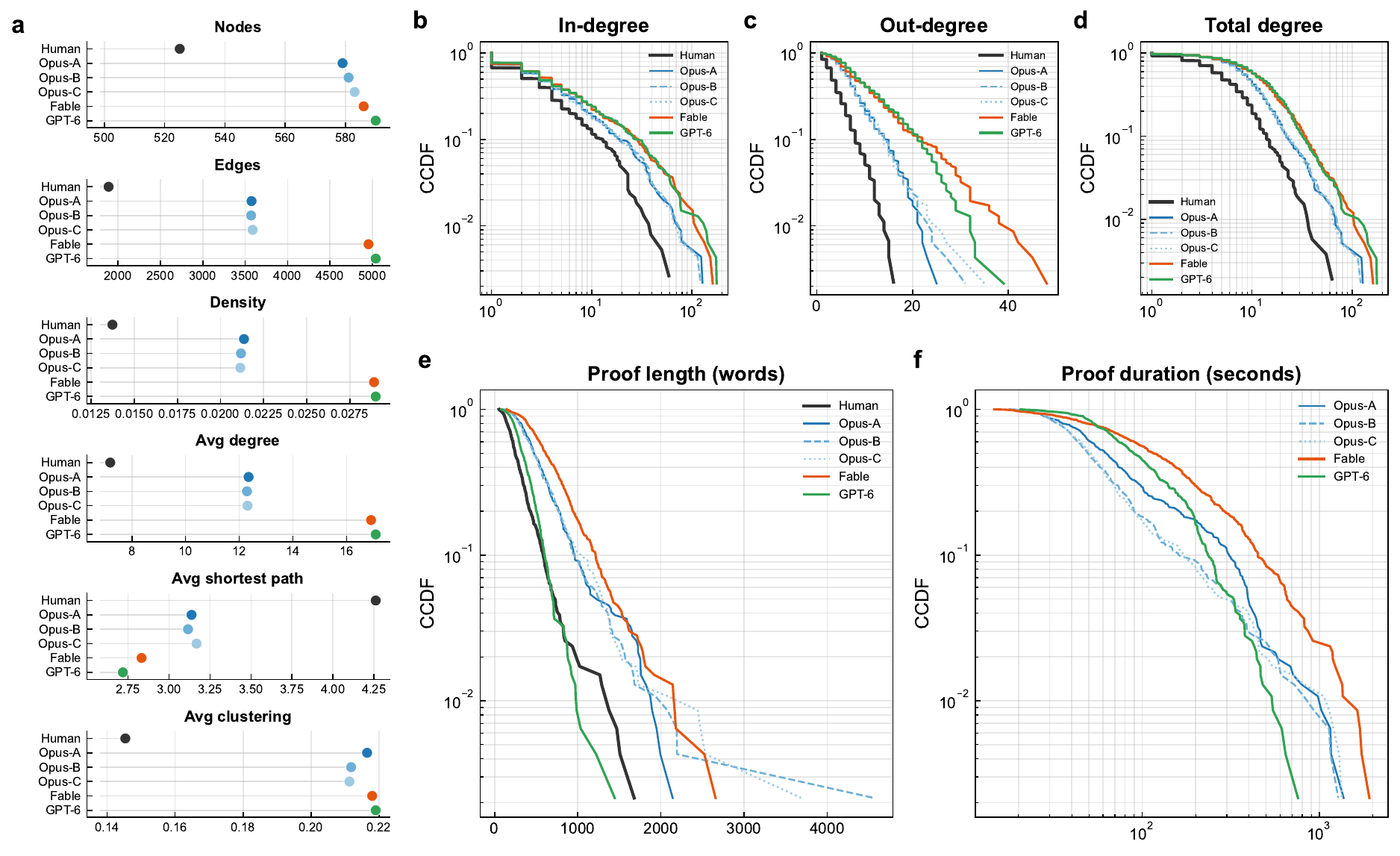}
\caption{\textbf{Descriptive statistics of proof citation networks and proof characteristics.}
{
\textbf{a} Six topological properties of the citation networks treated as undirected graphs: node count, edge count, density, average degree, average shortest path length (computed on the largest connected component), and average clustering coefficient.
Each dot represents one network; colors identify the six proof sources throughout the figure.
\textbf{b--d} Complementary cumulative distribution functions (CCDFs) of in-degree, out-degree, and total degree for all six networks.
In-degree and total degree are plotted on log-log axes; out-degree is plotted on a semi-logarithmic axis (linear x, logarithmic y).
\textbf{e} CCDF of proof length measured in words.
AI proof lengths are counted from the step-by-step proof text, excluding citation metadata; human proof lengths are counted from Heath's English translation.
Semi-logarithmic axes.
\textbf{f} CCDF of wall-clock proof duration for the five AI runs (no human counterpart), plotted on log-log axes.
}
}
\label{figure2}
\end{figure}

Before examining stylistic differences, we characterise the six proof citation networks and proof output characteristics at an aggregate level (Fig. \ref{figure2}).
All five AI networks are substantially denser than the human network (Fig. \ref{figure2}a).
The human network contains 525 nodes and 1,889 edges.
AI citation networks contain more nodes because the \textit{Elements} lists many definitions (132 in total) that are not explicitly cited in its own proofs, whereas AI proofs cite them during the proving process.
Claude Opus produces roughly twice as many edges (${\sim}$3,575 on average across three runs), while Fable and GPT-6 each exceed 4,900 edges---2.6 times the human count (1,889 edges).
This difference propagates through every topological summary: average degree rises from 7.2 (human) to 12.3 (Opus) and 17.0 (Fable/GPT-6); average shortest path length drops from 4.26 to 3.1 (Opus) and 2.7--2.8 (Fable/GPT-6); and average clustering coefficient increases from 0.145 to 0.21--0.22.

The in-degree and total degree distributions of all six networks exhibit heavy-tail-like characteristics (Fig. \ref{figure2}b, d); because AI networks have higher average degree, their CCDF curves are systematically shifted to the right.
By contrast, the out-degree distributions decay approximately exponentially (Fig. \ref{figure2}c).
This combination---exponential out-degree, power-law-like in-degree---is consistent with previous findings on human mathematical citation networks \supercite{liSocietalCitationsUndermine2025}, indicating that AI proof networks follow the same degree distribution patterns as their human counterpart without substantial change; the difference is that AI proofs cite more references on average.

For AI-generated content, we examine two features: proof text length (word count) and proof duration (seconds) (Fig. \ref{figure2}e, f).
AI proofs are consistently longer than Heath's text.
Human proofs average 314 words (median 248); Opus averages 568 words, Fable 702 words, and GPT-6 375 words.
Despite citing the most references per proposition (mean out-degree 8.5), GPT-6 produces the most concise AI proofs---each citation is accompanied by roughly 35 words of reasoning, compared with 74 words per citation in Opus proofs.
Wall-clock proof duration varies across models: Opus is fastest (median 50--63 s across three runs), GPT-6 is intermediate (median 90 s), and Fable is slowest (median 115 s, with a tail extending beyond 1,900 s).
The models thus occupy distinct positions along a verbosity-efficiency trade-off: GPT-6 is terse and moderately fast, Opus is moderately verbose and fast, and Fable is the most detailed but slowest.

\subsection*{AI has its own proving style}

\begin{figure}[ht]
  \centering
  \includegraphics[width=0.7\textwidth]{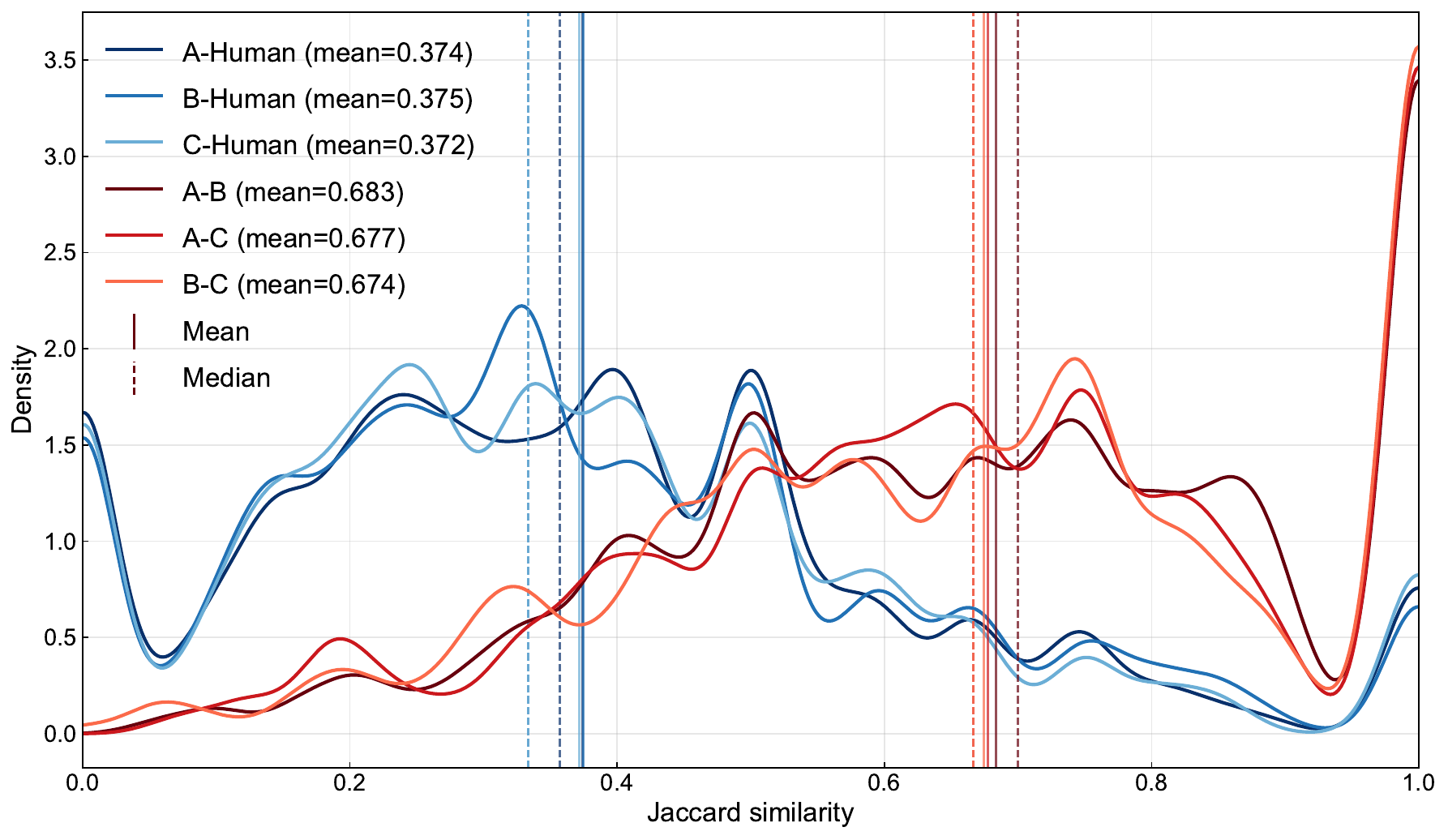}
\caption{\textbf{Pairwise Jaccard citation similarity across AI runs and human references.}
{
Kernel density estimates of the Jaccard similarity between citation sets, computed for each of the 465 propositions.
Blue curves show AI--human similarity (each run versus human references; mean $\approx 0.37$); red curves show intra-AI similarity (pairwise between runs; mean $\approx 0.68$). Vertical solid and dashed lines mark the group mean and median, respectively.
The clear separation between the two groups indicates that AI runs agree with each other substantially more than they agree with human references.
Cross-model results, including those for Fable 5.1 and GPT-6 Astra, are shown in Supplementary Tab. S10 and Supplementary Tab. S11.
}
}
\label{figure3}
\end{figure}

To determine whether AI develops its own citation style, we compute the Jaccard similarity between citation sets for each proposition across all runs, both AI-AI and AI-Human.
Fig. \ref{figure3} shows the distributions of pairwise Jaccard similarity among the three Opus runs and between each Opus run and the human reference.
The AI-Human distributions are shifted to the left, while the AI-AI distributions are shifted to the right; mean Jaccard nearly doubles from AI-Human to AI-AI, indicating that AI proofs are more similar to each other than to the human proofs.
Supplementary Tabs. S10 and S11 show the citation set similarity between the other models and Opus as well as the human reference, confirming the consistent pattern: AI-AI similarity is overall higher than AI-Human similarity.
In other words, AI develops its own citation style that departs from the human one.

To determine whether this style is stable, we examine the consistency of classification results---that is, whether AI proofs of the same proposition are always assigned to the same category (here we use the classification results after expert validation, which include the E class).
For example, across the three Opus runs, category P has a total count of 64, but these involve only 28 distinct propositions: 15 propositions are classified as P in all three runs, 6 in exactly two runs, and 7 in only one run.
That is, for many P-class propositions, AI repeatedly produces proofs whose citation sets match the human reference, maintaining stability.
We can then compute the 3-run overlap ratio (45 divided by 64) and the $\geq$2-run overlap ratio (57 divided by 64) to quantify this consistency.
A high ratio indicates strong consistency.
The same computation is applied to G- and D-class propositions, showing that for the same proposition, AI tends to stably produce the same type of proof (see Tab. \ref{tab2}).
However, for category E, no proposition is classified as E in all three Opus runs simultaneously, meaning that E-class outcomes are incidental rather than systematic.
In Supplementary Tab. S12, we show the consistency between the classifications of Fable 5.1 and GPT-6 Astra proofs and those of the three Opus runs---that is, whether the category assigned to a given proposition's proof by these two models agrees with Opus, and if so, with how many of the three runs.
The results show that approximately 57\% of propositions achieve three-run agreement and 14\% achieve two-run agreement.
These findings indicate that correct AI-generated proofs not only exhibit their own style, but that this style is stably produced rather than incidental.

\begin{table}[ht]
\centering
\caption{\textbf{Cross-run classification consistency.}}
\begin{threeparttable}
\begin{tabular}{lccccc}
\toprule
 & & \multicolumn{2}{c}{3-run overlap} & \multicolumn{2}{c}{$\geq$2-run overlap} \\
\cmidrule(lr){3-4} \cmidrule(lr){5-6}
Category & Total & Count & \% & Count & \% \\
\midrule
P &  64 &  45 & 70.3 &  57 & 89.1 \\
G & 934 & 723 & 77.4 & 873 & 93.5 \\
D & 371 & 201 & 54.2 & 303 & 81.7 \\
E &  26 &   0 &  0.0 &   8 & 30.8 \\
\midrule
\textbf{Overall} & \textbf{1395} & \textbf{969} & \textbf{69.5} & \textbf{1241} & \textbf{89.0} \\
\bottomrule
\end{tabular}
\begin{tablenotes}[flushleft]\small
\item \textbf{Note:} Total is the sum of each category's count across three independent runs. If a proposition is classified as P in all three runs, all three contribute to the 3-run overlap count; if classified as P in exactly two runs, two contribute to the $\geq$2-run overlap count. Percentages represent classification consistency.
\end{tablenotes}
\end{threeparttable}
\label{tab2}
\end{table}

\subsection*{Greater complexity, greater divergence}

\begin{figure}[htbp]
  \centering
  \includegraphics[width=\textwidth]{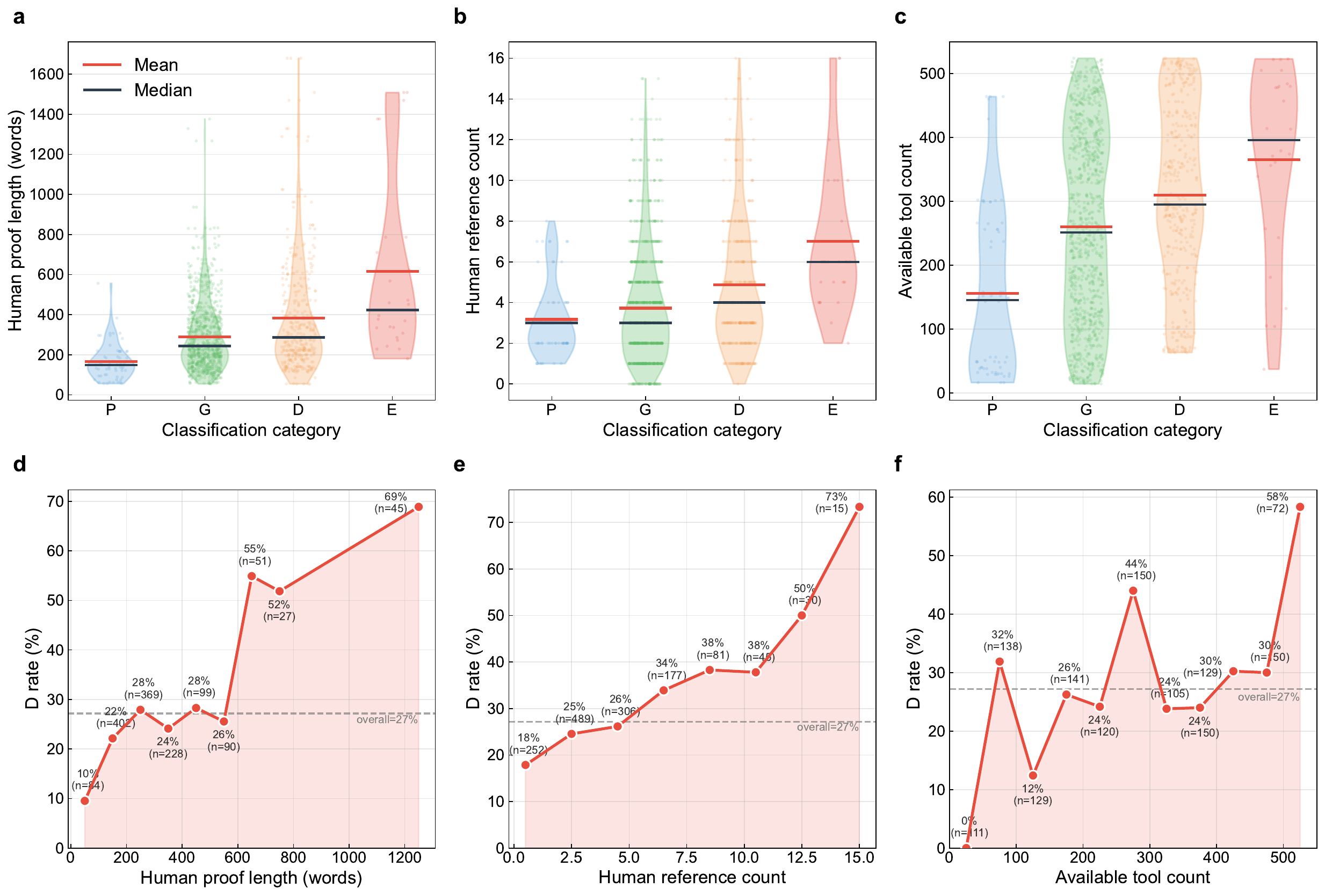}
\caption
{\textbf{Classification distribution and path-divergence rate across human proof complexity.}
{\textbf{a--c} Violin plots show the distribution of three human proof complexity proxies---proof length, reference count, and available tool count---across expert-verified classification categories (P, G, D, E; three independent runs pooled, $n = 1{,}395$).
Red and dark lines indicate the mean and median, respectively.
Categories with greater divergence from human proofs (D, E) tend to have higher complexity proxies than those with closer agreement (P, G), indicating that proof complexity is associated with AI--human divergence.
\textbf{d--f} Path-divergence (D) rate computed within binned intervals of each complexity proxy.
Propositions are grouped into bins of width 100 words, 2 references, and 50 available tools, respectively, with the final bin widened to capture the tail.
In each line plot, the horizontal coordinate of each point corresponds to the midpoint of its bin, and the vertical coordinate represents the proportion of D-classified propositions within that bin.
The dashed grey line marks the overall D~rate.
D~rate increases with all three proxies, indicating that AI proof paths diverge more frequently on complex propositions.
Results for the Fable 5.1 and GPT-6 Astra models are available in Supplementary Fig. S2 and Supplementary Fig. S3, respectively.
}
}
\label{figure4}
\end{figure}

Our classification (after expert validation), from P to G, D, and then to E, represents progressively greater AI-Human proof divergence.
In this section, we explore the sources of this divergence.
An intuitive hypothesis is that it may be related to the difficulty of the proposition itself.
To investigate this, we use three features of the proving process as proxies for proposition difficulty: Heath's proof length in words, Heath's reference count, and the number of citable resources available at the time of proving (postulates, common notions, all preceding definitions, and all preceding theorems).
Fig. \ref{figure4}a--c show the distribution of these proxies across the four classification categories (pooling the three Opus runs for a total of 1,395 propositions).
All three proxies increase from P to E: P-class propositions have the shortest human proofs (mean 153 words), the fewest human references (mean 3.0), and the fewest available tools (mean 264); D-class propositions are substantially more complex (mean 387 words, 5.1 references, 413 tools); and E-class propositions are the most complex of all (mean 755 words, 8.2 references, 482 tools).
This gradient indicates that divergence between AI and human proofs correlates with proposition difficulty---simpler propositions tend to produce citation-identical or granularity-level matches, while more complex ones are more likely to elicit genuinely different proof paths or errors.

Another perspective is the proportion of D-class proofs.
We bin propositions by each difficulty proxy and compute the ``D rate''---the number of propositions classified as D divided by the total number in that bin---within each bin (Fig. \ref{figure4}d--f).
The D rate rises with all three proxies: from 10\% to 69\% as human proof length increases, from 18\% to 73\% as human reference count increases, and from 0\% to 58\% as available tool count increases, against an overall baseline of 27\%.
The pattern is intuitive: a simple proposition with a short proof and few available references affords limited room for alternative reasoning, so AI and human proofs converge; a complex proposition with a long proof and many available references offers multiple viable proof paths, and AI is more likely to select a different one.

Supplementary Figs. S2 and S3 show the results of applying the same analysis to Fable 5.1 and GPT-6 Astra.
GPT-6 Astra exhibits patterns consistent with Opus.
Fable 5.1 also shows consistent patterns in the distribution of the three proxies across classification categories.
However, Fable 5.1's D rate results are not consistent with the other models, likely because the small number of D-class proofs in its results (34 out of 465) is insufficient to reveal a statistically reliable pattern.

\subsection*{The AI style: building from the ground up}

Having established through citation similarity that AI develops its own proving style, what this style actually looks like remains unclear.
Here, we characterise it by analysing citation structure, in particular by comparing AI-Only and Human-Only citations for the same propositions.

\begin{figure}[htbp]
  \centering
  \includegraphics[width=\textwidth]{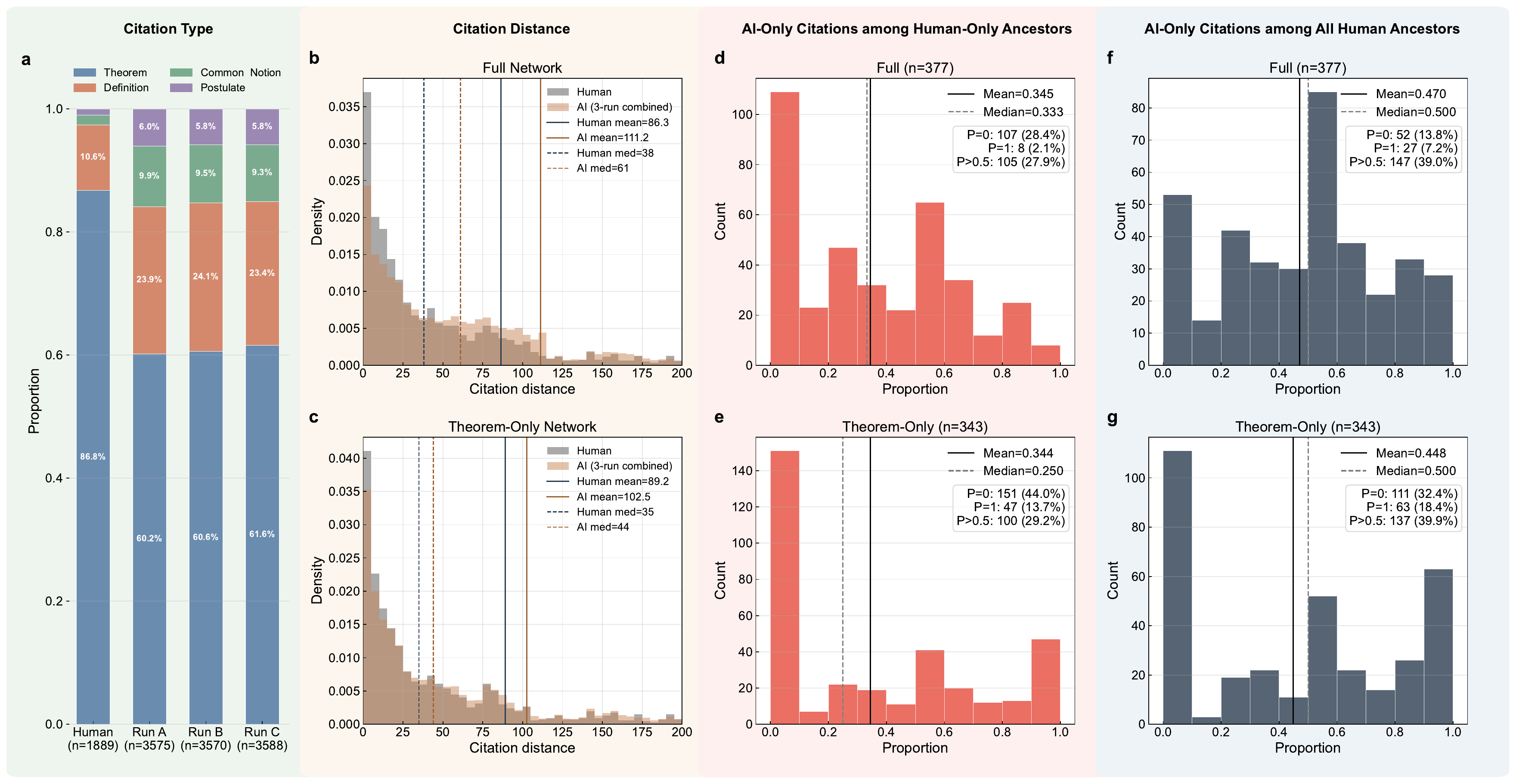}
\caption
{\textbf{Citation structure and human-ancestor overlap of AI proof paths.}
{\textbf{a} Proportion of citation types (theorem, definition, common notion, postulate) in human and AI networks.
AI proofs cite a lower share of theorems and a higher share of definitions than human proofs.
\textbf{b--c} Citation distance distributions for full and theorem-only networks.
Citation distance is the sequential index difference between a proposition and its cited reference.
AI distributions shift rightward (full network: AI mean = 111.2 vs Human mean = 86.3), indicating that AI proofs reach further back in the deductive chain.
\textbf{d--e} Distribution of the proportion of AI-only citations that appear among the ancestors of human-only citations (the transitive closure of the human-only citation set) for each D-class proposition.
\textbf{f--g} Distribution of the proportion of AI-only citations that appear among all human ancestors (the transitive closure of the full human citation set).
On average, each D-class proposition has 47\% of its AI-only citations falling within the human ancestor set, indicating that many AI path divergences represent recombinations of earlier underlying deductive material rather than direct citations of later mature propositions built on the same material.}
Similar results for the Fable 5.1 and GPT-6 Astra models are shown in Supplementary Fig. S4 and Supplementary Fig. S5, respectively.
}
\label{figure5}
\end{figure}

We first examine differences in citation type (Fig. \ref{figure5}a).
In the human network, 86.8\% of all citations point to theorems, with definitions, common notions, and postulates together accounting for only 13.2\%.
In the AI networks (Opus), theorems account for only ${\sim}$60\%, while definitions (${\sim}$24\%), common notions (${\sim}$10\%), and postulates (${\sim}$6\%) together make up ${\sim}$40\% of all citations.
Supplementary Figs. S4 and S5 show that for Fable and GPT-6, the proportion of theorem citations drops further to 59.6\% and 52.5\%, respectively.
This means that AI proofs rely more heavily on the three types of foundational elements.

The distribution of citation distance further reinforces this pattern.
Since each proposition in the Elements has a sequential number, we can determine the citation distance based on numbering.
For example, Proposition 4.4 cites Proposition 3.16, and the distance between them is 25 (Book III contains 37 propositions).
All postulates and common notions are assigned to the position of 1.1; definitions in each book are assigned to the position of the first proposition in that book.
Fig. \ref{figure5}b shows the distribution of citation distance for AI (Opus) and human proofs. The mean distance is 111.2 for AI versus 86.3 for human; the median is 61 versus 38.
This indicates that AI tends to cite more distant content.

However, a notable question is whether this tendency to cite more distant propositions is merely driven by AI's more frequent citation of foundational elements such as postulates and common notions, especially given that their positions are set at 1.1.
We therefore also examine citations among theorems alone.
We construct a theorem-only network in which nodes consist solely of propositions, excluding the three other types.
The citation distance distribution under this setting is shown in Fig. \ref{figure5}c.
The result is the same: AI's citation distances remain longer than those of the human proof.
Supplementary Figs. S4 and S5 show that the results for Fable and GPT-6 are also consistent with Opus in terms of citation distance.
Moreover, we also computed citation distance in terms of topological generations\supercite{healy2013hierarchical,liSocietalCitationsUndermine2025}, and the results are consistent (Supplementary Tab. S13).
These results indicate that AI not only cites foundational postulates, common notions, and definitions more frequently, but also tends to cite earlier, more distant content even among proposition-to-proposition citations.

We have already seen AI's tendency to build arguments from foundational elements rather than directly relying on later, more established propositions.
We further examine D-class propositions to substantiate this point.
D-class propositions are those where AI and human proof paths genuinely diverge.
If AI builds arguments from the ground up, we want to understand the nature of this divergence---is AI merely ``reinventing the wheel'', ostensibly constructing from foundations but in reality simply retracing the steps required to derive those later, established propositions?
We investigate this by checking whether a proposition's AI-Only citations appear among the ancestors of the human proof's citations.
A proposition's AI-Only citations are references cited only by the AI proof, excluding those cited only by the human proof or those cited by both AI and human.
The ancestor nodes of a proposition are its cited nodes, the cited nodes of those cited nodes, and so on, tracing back to the most foundational elements---representing the full set of upstream dependencies of that proposition.
For each D-class proposition, we compute the proportion of its AI-Only citations that appear among the ancestors of the human proof's citations.

Fig. \ref{figure5}f shows the distribution of this proportion.
The mean of this proportion across all D-class propositions is 47.0\%; when restricted to ancestors of Human-Only citations, the mean proportion is 34.5\% (Fig. \ref{figure5}d).
In the theorem-only network, the mean proportion is 44.8\% among all human ancestors (Fig. \ref{figure5}g) and 34.4\% among Human-Only ancestors (Fig. \ref{figure5}e).
For Fable 5.1 (Supplementary Fig. S4), the mean proportions are 52.7\% and 42.7\% in the full network (All Human and Human-Only ancestors, respectively) and 55.5\% and 49.1\% in the theorem-only network.
For GPT-6 Astra (Supplementary Fig. S5), the corresponding values are 48.2\% and 39.0\% in the full network and 45.5\% and 36.7\% in the theorem-only network.
These proportions indicate that AI's extra citations are not unrelated to the human proof.
When roughly half of a proposition's AI-Only citations can be traced to the upstream dependencies of the human proof, it suggests that AI and human are often proving the same result through the same underlying logical structure---but entering the dependency chain at different levels.
The human proof cites mature, high-level propositions that encapsulate intermediate results; the AI proof independently cites the more foundational elements from which those high-level propositions are derived.

\begin{table}[ht]
\centering
\begin{threeparttable}
\caption{\textbf{D-class AI-Only and Human-Only subgraph statistics}}
\label{tab3}
\begin{tabular}{l cccc}
\toprule
 & \multicolumn{2}{c}{Full Network} & \multicolumn{2}{c}{Theorem-Only} \\
\cmidrule(lr){2-3} \cmidrule(lr){4-5}
Metric & AI-Only & Human-Only & AI-Only & Human-Only \\
\midrule
\addlinespace[8pt]
\textbf{All D-class pairs} \\
\addlinespace[4pt]
Sample size              & 377    & 352    & 343    & 351    \\
Mean node count          & 6.33   & 2.87   & 3.76   & 2.68   \\
Has internal edges       & 76.9\% & 29.5\% & 51.6\% & 27.4\% \\
Longest path $\geq$ 2    & 28.4\% & 7.4\%  & 17.5\% & 6.6\%  \\
Mean longest path        & 1.19   & 0.37   & 0.75   & 0.34   \\
Mean largest component   & 3.94   & 1.53   & 2.34   & 1.46   \\
Largest component ratio  & 0.608  & 0.650  & 0.686  & 0.669  \\
\addlinespace[8pt]
\textbf{Subgraphs with $\geq$ 3 nodes} \\
\addlinespace[4pt]
Sample size              & 339    & 148    & 205    & 137    \\
Mean node count          & 6.85   & 4.74   & 5.32   & 4.56   \\
Has internal edges       & 83.5\% & 57.4\% & 78.0\% & 54.7\% \\
Mean longest path        & 1.30   & 0.76   & 1.17   & 0.72   \\
Mean largest component   & 4.25   & 2.14   & 3.16   & 2.02   \\
Largest component ratio  & 0.594  & 0.456  & 0.580  & 0.447  \\

\bottomrule
\end{tabular}
\begin{tablenotes}
\small
\item \textbf{Note:} Largest component ratio = size of the largest connected component divided by the total number of nodes in the subgraph, averaged across all qualifying propositions.
Similar results for the Fable 5.1 and GPT-6 Astra models are shown in Supplementary Tab. S14 and Supplementary Tab. S15, respectively.
\end{tablenotes}
\end{threeparttable}
\end{table}

Furthermore, we find that AI-Only citations exhibit stronger internal connectivity. For each D-class proposition, we compare the induced subgraph formed by AI-Only citation nodes with the induced subgraph formed by Human-Only citation nodes  (see Tab. \ref{tab3}).
AI-Only subgraphs are substantially larger (mean 6.33 nodes versus 2.87 for Human-Only in the full network) and, critically, far more internally connected: 76.9\% of AI-Only subgraphs contain at least one internal edge, compared with only 29.5\% for Human-Only.
The mean longest path within AI-Only subgraphs is 1.19 versus 0.37 for Human-Only.
These differences persist in the theorem-only network (51.6\% versus 27.4\% with internal edges; mean longest path 0.75 versus 0.34).
Since Human-Only subgraphs have fewer nodes on average, they are inherently less likely to form connected structures.
To control for this, we also restrict the comparison to subgraphs with at least three nodes (see Tab. \ref{tab3}); the result is the same---AI-Only subgraphs remain more connected (83.5\% versus 57.4\% with internal edges in the full network).
Similar results for Fable 5.1 and GPT-6 Astra are shown in Supplementary Tab. S14 and Supplementary Tab. S15, respectively.
In addition, we binned subgraphs by node count and computed the proportion with internal edges within each bin; the conclusion remains consistent after controlling for subgraph size (Supplementary Fig.~S6).
These results reveal a consistent structural characteristic of AI proving behaviour.
Human-Only citations typically point to high-level, established propositions that are unlikely to have direct dependency relationships with one another.
By contrast, AI-Only citations include more foundational propositions, which are more likely to form direct dependencies among themselves.
This is consistent with the ancestor analysis presented earlier.
In short, these results collectively indicate that AI tends to build its proofs from the ground up.

Finally, to test whether this tendency can be steered by prompt design, we conducted supplementary experiments with four prompt conditions applied to GPT-6 Astra on all 465 propositions (see Supplementary Information).
An explicit anti-grounding instruction---directing the model to prefer mature, previously proved propositions---had negligible effect (Cohen's $d = 0.047$).
Replacing the inline resource listing with an external table that made earlier-book theorem statements accessible produced a small improvement in Jaccard similarity ($d = 0.191$--$0.229$), but the AI's foundation citation ratio (42--48\%) remained far above the human level.
These results suggest that the ground-up tendency reflects an intrinsic property of how the model constructs proofs, rather than a prompt-engineering artefact.

A concrete example is Proposition~5.18, where the Jaccard similarity is zero.
The human proof is an indirect argument based on reductio ad absurdum, citing three high-level theorems (5.17, 5.14, 5.11); the AI proof returns to the equimultiple definition of proportion (D5.5) and constructs a direct verification from foundational elements.
The two proofs reach the same mathematical conclusion through different levels of the deductive hierarchy (see Supplementary Text 2 for a detailed dependency analysis).

\section*{Discussion}

Our results provide evidence that AI develops its own proving style, one that is stable across independent runs and models.
This style can be summarised as ``building from the ground up'': AI tends to construct logical chains on its own from foundational elements rather than directly citing later, more established results.
This tendency is not affected by changes in prompting, including changes in how citable resources are presented to the AI.
In addition, we find that the more complex a proposition is, the more likely AI is to produce a proof that differs from the human one.

A plausible explanation lies in how AI encounters mathematical knowledge.
Human mathematicians directly cite mature results from the existing body of work---to some extent, this is precisely why mathematicians establish such results: to conserve cognitive resources and facilitate the proving of subsequent propositions.
In our experimental setup, AI treats all citable elements on equal footing, with no implicit hierarchy among them.
AI can adopt a direct strategy to achieve the goal of proving a proposition, without having to account for additional constraints or develop a preference for citing higher-level results.
In doing so, it may explore proving routes that differ markedly from those taken by human mathematicians.

The regularity we identify may also help explain some of the observed manifestations.
``Building from the ground up'' may contribute to the verbosity of AI-generated proofs and their tendency to dwell repeatedly on local details, as reconstructing logical chains from foundational elements can require more intermediate steps and explicit justifications than invoking established results.
It may also partly explain the preference for procedural approaches, as AI tends to construct proofs through explicit sequences of operations rather than rely on higher-level mathematical results, an approach that requires a degree of mathematical insight.
This progression from an underlying tendency to its observable manifestations ultimately means that the correctness of an AI-generated proof does not guarantee its clarity, cognitive accessibility, or structural quality\supercite{petrovNotAllProofs2026,fuProofRefactorRefactoringGenerated2026,baudart2026library}.

This distinction is also reflected in recent responses from the mathematical community: the Leiden Declaration on Artificial Intelligence and Mathematics identifies ``certainty and understanding'' as the dual purpose of mathematical proof\supercite{alper2026leidendeclarationartificialintelligence}; the twenty-five Fields Medalists who signed an open letter on AI in mathematics expressed a similar concern\supercite{tao2026misalignment}.
Existing research on the readability of traditional machine-generated proofs has examined how the organisation of proof steps and the distance between premises and the steps they justify relate to legibility\supercite{pakImprovingLegibilityFormal2015,quaresmaMeasuringReadabilityGeometric2023}.
These studies motivate us to consider whether the structural properties of AI-generated proofs that we observe are also associated with their interpretability.
If a proof repeatedly reaches back to distant propositions or foundational elements---as reflected in the longer citation distances we observe (AI mean 111.2 versus human 86.3)---readability is expected to decline.
Moreover, the higher internal connectivity of AI-Only subgraphs (76.9\% containing internal edges, compared with 29.5\% for Human-Only) means that AI's additional citations are not independent references but are interconnected by dependencies.
Readers must track not only individual citations but also the relationships among them, potentially incurring additional cognitive cost.
Together, these findings suggest a possible structural explanation for why AI-generated proofs may be less intuitive to readers accustomed to human mathematical conventions.

Although our experiments show that changes in prompting do not significantly alter this proving style, other means of influencing AI's proof construction cannot be ruled out.
These may include fine-tuning models on human-written proofs, incorporating explicit citation preferences into training objectives, or modifying the proof-generation harness to encourage proof paths that make greater use of established mathematical results.
How to successfully steer AI toward proof paths that better align with human mathematical conventions remains to be determined through further experiments.
Conversely, once mathematicians recognise that AI proofs tend to reconstruct from foundations rather than cite established shortcuts, they can anticipate this pattern when reading AI-generated proofs.
The concern that AI proofs are ``not written for humans'' may be partly alleviated by recognizing that, at least in our experimental setting, AI-generated proofs follow a different but partly predictable convention for selecting mathematical resources.
The effectiveness of future human--AI collaboration in mathematics will benefit from this kind of characterisation of conventions of each side and the targeted improvements it enables.

Several limitations should be noted.
First, the present study is conducted within the setting of Euclid's ``Elements''; extending this approach to more complex domains of modern mathematics would help determine whether the patterns we observe are generalisable.
Second, our comparison relies on Heath's proof text as the human baseline; incorporating proofs by other mathematicians would offer a fuller picture of the variation within human proving itself.
Third, the present study focuses on proving style in natural language; formal proofs are not addressed here.
Looking ahead, two directions appear particularly promising.
One is to test whether the ``building from the ground up'' tendency can be steered and whether doing so improves the interpretability of the resulting proofs without sacrificing correctness.
The other is to extend this style of comparison to formal proof environments.
As mathematicians increasingly embrace the formalisation of mathematical proofs, whether AI and human proofs exhibit stylistic differences in formal languages warrants closer investigation.

\clearpage
\section*{Methods}
\subsection*{Data}

The dataset is drawn from a structured version of Heath's classical English translation of Euclid's \textit{Elements}, hosted on the Wolfram Data Repository (in JSON format)\supercite{wolframStructureEuclidsElements2020}.
The full text comprises 5 postulates, 5 common notions, 132 definitions, and 465 propositions across 13 books, totalling 607 statements.
The human proof text was extracted from the ``Proof'' field of each proposition in the dataset.
The human citation record was extracted from the ``References'' field of each proposition, which lists all premises invoked in each proof.
Citation identifiers follow a uniform coding scheme: postulates are labelled P1-P5, common notions CN1-CN5, definitions D\textit{book}.\textit{number} (e.g. D1.15, D10.2.1), and propositions \textit{book}.\textit{number} (e.g. 1.47, 9.36).
Parsing ultimately produced three files: a table of all 607 statements, a table recording the citation list for each of the 465 propositions, and a table recording the human proof text for each of the 465 propositions.

\subsection*{AI proof generation}

We used three large language models to independently reprove all 465 propositions: Claude Opus 4.6 (three independent runs, denoted RunA, RunB, RunC), Claude Fable 5.1 (one run), and GPT-6 Astra (one run).
The reasoning effort of all models was set to the second-highest available tier: ``high'' for Claude Opus 4.6, and ``xhigh'' for both Claude Fable 5.1 and GPT-6 Astra.
Proofs were generated one book at a time: for each book, a workflow spawned one sub-agent per proposition, proving them in parallel.
The workflow controlled the prompt fed to each sub-agent: 
For each proposition \textit{N}.\textit{k}, the sub-agent received the proposition statement together with a filtered inventory of citable resources: all postulates, common notions, and definitions, plus all propositions from books 1 through \textit{N}$-$1 and propositions 1 through \textit{k}$-$1 within book \textit{N}.
The sub-agent was instructed to produce a step-by-step proof citing only elements from this list; it then output the proof text and a USED list recording every element actually invoked during its reasoning.

After all propositions in a book were completed, the workflow executed a post-processing step that computed the Jaccard similarity for each proposition by directly comparing the AI and human citation sets.
For each proposition~$i$, let $A_i$ and $B_i$ denote the citation sets of two proof sources.
The Jaccard similarity coefficient is
\begin{equation}
J(A_i, B_i) = \frac{|A_i \cap B_i|}{|A_i \cup B_i|}.
\end{equation}
The post-processing step then compiled the proof text for each proposition in the current book---together with the AI citation set, the human citation set, and the computed Jaccard similarity---into a single Markdown file, and output a table recording the AI and human citation sets.
The complete project files for AI proof generation, executed through Claude Code, are available at \href{https://github.com/ffzdshz/AI_math_style}{GitHub}.

\subsection*{AI check}

Each AI proof underwent two stages of automated checking: classification and, for geometric propositions, a geometric consistency audit.

\noindent\textbf{Classification.} A separate AI judge classified every AI proof into one of the four categories (P, G, D, E) by following a structured decision protocol.
Classification was performed through a parallel workflow, with a dedicated sub-agent classifying each proposition independently.
Each sub-agent received the AI proof text, the human proof text, and a table comparing the two citation sets.
The protocol proceeds in fixed order, stopping at the first definitive answer.

First, a forward-dependency gate (automated and binding) checks whether the AI proof cites any proposition not yet established at the time of proving.
If any forward citation is found, the proof is immediately classified as E.

Second, a perfect-match check compares the AI and human citation sets.
If the two sets are identical, the proof is classified as P; no further analysis is needed.
Next, the sub-agent examines several questions in sequence.

Question 1 asks whether the AI proof is logically complete, and comprises two sub-questions.
Question 1(a) checks case coverage: for propositions that require multiple cases, each required case must be addressed by a dedicated argument or by a single general argument that demonstrably covers it; a genuinely absent case triggers E.
Question~1(b) checks whether the proof's conclusion is validly established, under a hardened standard: the default presumption is that the proof is valid, and E is assigned only when a defect falls under a closed list of three error modes---unsupported assertion (a claim has no supporting step and no citation that could license it), circular or vacuous citation (the only justification merely names or defines the thing to be proved), and invalid inference (the cited proposition does not entail the claimed result).
A conservative rule governs borderline cases: when a step is terse but supported, or compressed but sound, the verdict is not E.
An explicit audit record must be completed before any E assignment; if the record cannot be completed convincingly, the verdict defaults to not E.

Question 2 asks whether both proofs argue through the same key intermediate conclusions---the indispensable facts forming each proof's logical backbone.
Constructive sub-steps (drawing a line, placing a point) are not key conclusions.
If both proofs share the same backbone (Case A), the difference is one of citation depth (G).
If the backbones differ (Case B), the protocol proceeds to Question 3.

Question~3 asks whether the AI's different path is valid.
Positional validity (no forward dependencies) has already been established by the gate.
The remaining check is logical completeness: the AI's reasoning chain must run from premises to conclusion.
If the chain is complete, the proof is classified as D.
If a genuine logical gap is found under the hardened Question 1(b) standard, the proof is classified as E.
If the audit record cannot be completed convincingly, the verdict defaults to D.

\noindent\textbf{Geometric consistency audit.} All D-class proofs underwent a separate check for spatial configuration contradictions---errors in which individual proof steps each appear logically sound but their combined spatial assumptions are mutually inconsistent.
The audit examined six categories of spatial claims: point position consistency, ray and line ordering, angle decomposition validity, triangle interior/exterior arguments, circle containment, and (for solid geometry) plane membership and dihedral relationships.
Each proof received a verdict of PASS, FAIL, or SKIP (for purely algebraic propositions with no spatial claims).
Proofs judged FAIL were reclassified as E.

After the classification and geometric consistency audit were complete, we obtained the classification results reported as ``After AI check'' in Tab. \ref{tab1}.
The complete project files for AI classification and geometric consistency audit are available at \href{https://github.com/ffzdshz/AI_math_style}{GitHub}.

\subsection*{Expert validation and re-prove}

Two experts with mathematical training---the second and third authors---reviewed the results of the AI check.
The review covered two tasks: validating all identified errors and checking a random sample of D-class proofs.
For error validation, the experts reviewed all 29 items flagged as errors across the three Opus runs: 18 E-class proofs from the classification stage and 11 D-class proofs reclassified as~E after receiving FAIL verdicts in the geometric consistency audit.
For each item, the experts received the AI proof text and the corresponding error analysis, and assessed whether the identified error was genuine.
This review resulted in three corrections, all involving Propositions~9.9 and~9.10.
The statements of these two propositions concern only square numbers; Heath's proof additionally addresses the cube case, but this goes beyond what the proposition statement requires.
The AI proved exactly what the statement asked for.
By comparing the AI proofs against Heath's more extensive proof, the AI judge had incorrectly treated the omission of the cube case as a missing-case error.
The experts determined that the AI's proofs were correct: two were reclassified as~G and one as~D.
All remaining 26 errors were confirmed as genuine. The classification after expert validation is reported as ``After expert validation'' in Tab.~\ref{tab1}.

To verify the accuracy of the D classification, the experts also reviewed a random sample of 18 D-class proofs drawn from the 371 D-class proofs.
For each item, the experts received both the AI and human proof texts together with the classification report, and judged whether the proof constituted a genuine path difference rather than a granularity difference~(G) or an error~(E).
All 18 were confirmed as genuine path differences, although four were noted as borderline and ambiguous cases.
The complete expert validation reports are available at \href{https://github.com/ffzdshz/AI_math_style}{GitHub}.

Every confirmed error (26 proof--run pairs across the three runs) was then re-proved by the AI under the same constraints.
The model received the proposition statement, all citable elements, and a summary of the error to be avoided, but not the human proof text.
All 26 errors were successfully resolved: the re-proved proofs were classified as~G (18 cases) or~D (8 cases).
The final classification (Tab.~\ref{tab1}, ``After re-prove'') contains zero E-class proofs in all runs.

\subsection*{Network analysis}

\noindent\textbf{Citation networks.}
For each proof source (three Opus runs, Fable, GPT-6, and Heath), we constructed a directed citation network in which nodes represent all 465 propositions together with all definitions, common notions, and postulates cited by at least one proof, and directed edges point from each proposition to the elements it cites.
We also constructed theorem-only networks restricted to proposition-to-proposition citations, excluding definitions, common notions, and postulates.

\noindent\textbf{Citation distance.}
Each proposition in the \textit{Elements} has a sequential index determined by its position in the text.
The citation distance between a citing proposition and a cited element is their index difference.
All postulates and common notions are assigned the index of Proposition~I.1; definitions in each book are assigned the index of the first proposition in that book.
For example, Proposition~4.4 cites Proposition~3.16, and the distance between them is 25 (Book~III contains 37 propositions).

Furthermore, considering that the sequential numbering in the \textit{Elements} may simply reflect Euclid's personal arrangement, we also computed citation distances using an alternative measure based on topological generations.
A mathematical theorem network is a directed acyclic graph (DAG), in which each node can be assigned a topological generation $g$---a nonnegative integer indicating the layer at which the node is located in the overall network structure~\supercite{healy2013hierarchical,liSocietalCitationsUndermine2025}.
Formally, let $S(i)$ denote the set of successor nodes of node $i$ (in citation networks, directed links point from newer nodes to older nodes they cite, so a cited predecessor is a graph-theoretic successor).
The topological generation of node $i$ is defined as
\begin{equation}
g(i) =
\begin{cases}
0 & \text{if } |S(i)| = 0, \\
\max\{g(j) : j \in S(i)\} + 1 & \text{otherwise},
\end{cases}
\end{equation}
where $|\cdot|$ denotes the set cardinality.
Nodes with no outgoing citations (axioms, postulates, and definitions) belong to generation~0; all other theorems are assigned to the earliest possible generation, accumulating layer by layer from these foundations.
The citation distance between two nodes is then measured as the absolute difference in their topological generations.
The results are presented in Supplementary Tab.~S13.

\noindent\textbf{Ancestor analysis.}
To investigate the structural relationship between AI-only and human-only citations, we performed an ancestor analysis on the human citation network.
In a directed acyclic graph $G = (V, E)$, node $u$ is an \emph{ancestor} of node $v$ if there exists a directed path from $v$ to $u$~\supercite{BangJensen2009}.
The set of all ancestors of $v$ is denoted as
\begin{equation}
A(v) = \{u \in V : \text{there exists a directed path from } v \text{ to } u\}.
\end{equation}
In our citation networks, where a directed edge $(i, j)$ indicates that proposition $i$ cites proposition $j$, the ancestors of a node $v$ are all propositions transitively reachable by following outgoing citation links from $v$---that is, the complete set of foundational results upon which $v$ ultimately depends.

For each D-class proposition, we identified the AI-only citation set $C_{\mathrm{AI}}$ and the human-only citation set $C_{\mathrm{H}}$, defined as $C_{\mathrm{AI}} = C_{\mathrm{AI}}^{\mathrm{full}} \setminus C_{\mathrm{H}}^{\mathrm{full}}$ and $C_{\mathrm{H}} = C_{\mathrm{H}}^{\mathrm{full}} \setminus C_{\mathrm{AI}}^{\mathrm{full}}$, where $C_{\mathrm{AI}}^{\mathrm{full}}$ and $C_{\mathrm{H}}^{\mathrm{full}}$ are the complete citation sets of the AI and human proofs, respectively.
We then computed the ancestor set of the human-only citations in the human citation network $G_{\mathrm{H}}$:
\begin{equation}
A(C_{\mathrm{H}}) = \bigcup_{v \in C_{\mathrm{H}}} A_{G_{\mathrm{H}}}(v).
\end{equation}
The ancestor proportion for each proposition is defined as the fraction of AI-only citations that fall within this ancestor set:
\begin{equation}
P = \frac{|C_{\mathrm{AI}} \cap A(C_{\mathrm{H}})|}{|C_{\mathrm{AI}}|}.
\end{equation}
The proportion of AI-only citations that are ancestors of all human citations, $P' = |C_{\mathrm{AI}} \cap A(C_{\mathrm{H}}^{\mathrm{full}})|~/~|C_{\mathrm{AI}}|$, is defined analogously.
High values of $P$ and $P'$ indicate that the AI's distinctive citations are not arbitrary alternatives but share a deeper common foundation with the human citations.
This is consistent with a ``building from the ground up'' proving strategy.

\noindent\textbf{Subgraph analysis.}
To examine the internal dependency structure among the distinctive citations of each proof strategy, we analysed the induced subgraphs\supercite{Diestel2017} of AI-only and human-only citation sets.
For each D-class proposition, we extracted the AI-only set $C_{\mathrm{AI}}$ and the human-only set $C_{\mathrm{H}}$, and constructed their induced subgraphs $G[C_{\mathrm{AI}}]$ and $G[C_{\mathrm{H}}]$ in the respective citation networks.
For each subgraph, we measured the node count, the proportion containing at least one internal edge, the mean longest directed path, and the size and ratio of the largest connected component.
Results were reported both for all subgraphs and for those with three or more nodes; we further binned subgraphs by node count to confirm that the connectivity difference persists after controlling for subgraph size (Supplementary Fig.~S6).

\section*{Data availability}

The structured dataset of Euclid's \textit{Elements} used in this study is publicly available on the \href{https://datarepository.wolframcloud.com/resources/Structure-of-Euclids-Elements}{Wolfram Data Repository}.
All AI-generated proofs, citation comparison tables, classification results, expert validation reports, citation networks, and source data for all figures and tables are available on \href{https://github.com/ffzdshz/AI_math_style}{GitHub}.

\section*{Code availability}

All project files executed through Claude Code are available on \href{https://github.com/ffzdshz/AI_math_style}{GitHub}.
Code for statistical analysis and figure generation is also available in the same repository.

\printbibliography

\section*{Acknowledgements}

We thank Yunfei Zhang of the School of Mathematics and Statistics, Qingdao University, for helpful discussions on the research design.

\section*{Author contributions}

X.L. conceptualized the study, developed the methodology, performed formal analysis and investigation, curated the data, wrote the original draft, and created the visualizations.
Z.Z. contributed to conceptualization, performed validation, provided resources, and reviewed and edited the manuscript.
Q.C. performed validation, provided resources, and reviewed and edited the manuscript.
Y.W. performed investigation and reviewed and edited the manuscript.
A.Z. contributed to conceptualization and methodology, reviewed and edited the manuscript, and supervised the project.
Y.F. contributed to conceptualization and methodology, reviewed and edited the manuscript, supervised the project, administered the project, and acquired funding.

\section*{Competing interests}

The authors declare no competing interests.

\end{document}